\documentclass[10pt,twocolumn,letterpaper]{article}

\usepackage{cvpr}              

\usepackage[dvipsnames]{xcolor}

\usepackage[accsupp]{axessibility}
\usepackage{pifont}
\usepackage{graphicx}
\usepackage{amsmath}
\usepackage{amssymb}
\usepackage{booktabs}
\usepackage{tabularx}
\usepackage{multicol}
\usepackage{multirow}
\usepackage{tablefootnote}
\usepackage{bm}
\usepackage{ulem}
\usepackage{threeparttable}

\definecolor{cvprblue}{rgb}{0.21,0.49,0.74}
\usepackage[pagebackref,breaklinks,colorlinks,citecolor=cvprblue]{hyperref}

\def\confName{CVPR}
\def\confYear{2024}

\title{\vspace{-10pt}
Instance-Adaptive and Geometric-Aware Keypoint Learning \\
for Category-Level 6D Object Pose Estimation}

\author{
    \quad Xiao Lin$^{1}$
    \quad Wenfei Yang$^{1,2}$
    \quad Yuan Gao$^{1}$
    \quad Tianzhu Zhang$^{1,}$\footnotemark[1]
    \\
    $^1$University of Science and Technology of China
    \\
    $^2$Jianghuai Advance Technology Center, Hefei, 230000, China
    \\
    {\tt\small \{llinxiao,wazs98\}@mail.ustc.edu.cn \quad \{yangwf,tzzhang\}@ustc.edu.cn }
}

\begin{document}
\maketitle
\renewcommand{\thefootnote}{\fnsymbol{footnote}}
\footnotetext[1]{Corresponding author.} 
\begin{abstract}
Category-level 6D object pose estimation aims to estimate the rotation, translation and size of unseen instances within specific categories. In this area, dense correspondence-based methods have achieved leading performance. However, they do not explicitly consider the local and global geometric information of different instances, resulting in poor generalization ability to unseen instances with significant shape variations.
To deal with this problem, we propose a novel Instance-\textbf{A}daptive and \textbf{G}eometric-Aware Keypoint Learning method for category-level 6D object pose estimation (AG-Pose), which includes two key designs: 
(1) The first design is an Instance-Adaptive Keypoint Detection module, which can adaptively detect a set of sparse keypoints for various instances to represent their geometric structures. (2) The second design is a Geometric-Aware Feature Aggregation module, which can efficiently integrate the local and global geometric information into keypoint features. These two modules can work together to establish robust keypoint-level correspondences for unseen instances, thus enhancing the generalization ability of the model.Experimental results on CAMERA25 and REAL275 datasets show that the proposed AG-Pose outperforms state-of-the-art methods by a large margin without category-specific shape priors. Code will be released at \url{https://github.com/Leeiieeo/AG-Pose}
\end{abstract}

\section{Introduction}
\label{sec:intro}
The 6D object pose estimation task aims to predict the rotation, translation 
and size of objects with 2D/3D observations.
Due to its great potential in many real-world applications 
such as robotic manipulation \cite{robotmani2, robotmani3, robotmani5}, 
augmented reality \cite{ar1, ar3} and 
autonomous driving \cite{ad1, ad2, ad3},
this task has been attracting increasing attention 
in the research community. 
While many previous methods 
\cite{He2020pvn3d, peng2019pvnet, Wang2021GDRN, He2021fffb6d, liu2022gdrnpp_bop, wu2022vote} 
have achieved significant performance on instance-level 6D object pose estimation, 
their reliance on instance-level CAD models hinders their generalization to novel instances. 
To deal with this problem, 
category-level 6D object pose estimation has been introduced in \cite{Wang2019nocs},
which aims to estimate the poses of unseen instances 
within specific categories without using their CAD models.

\begin{figure}[t]
\begin{center}
\includegraphics[width=1\linewidth]{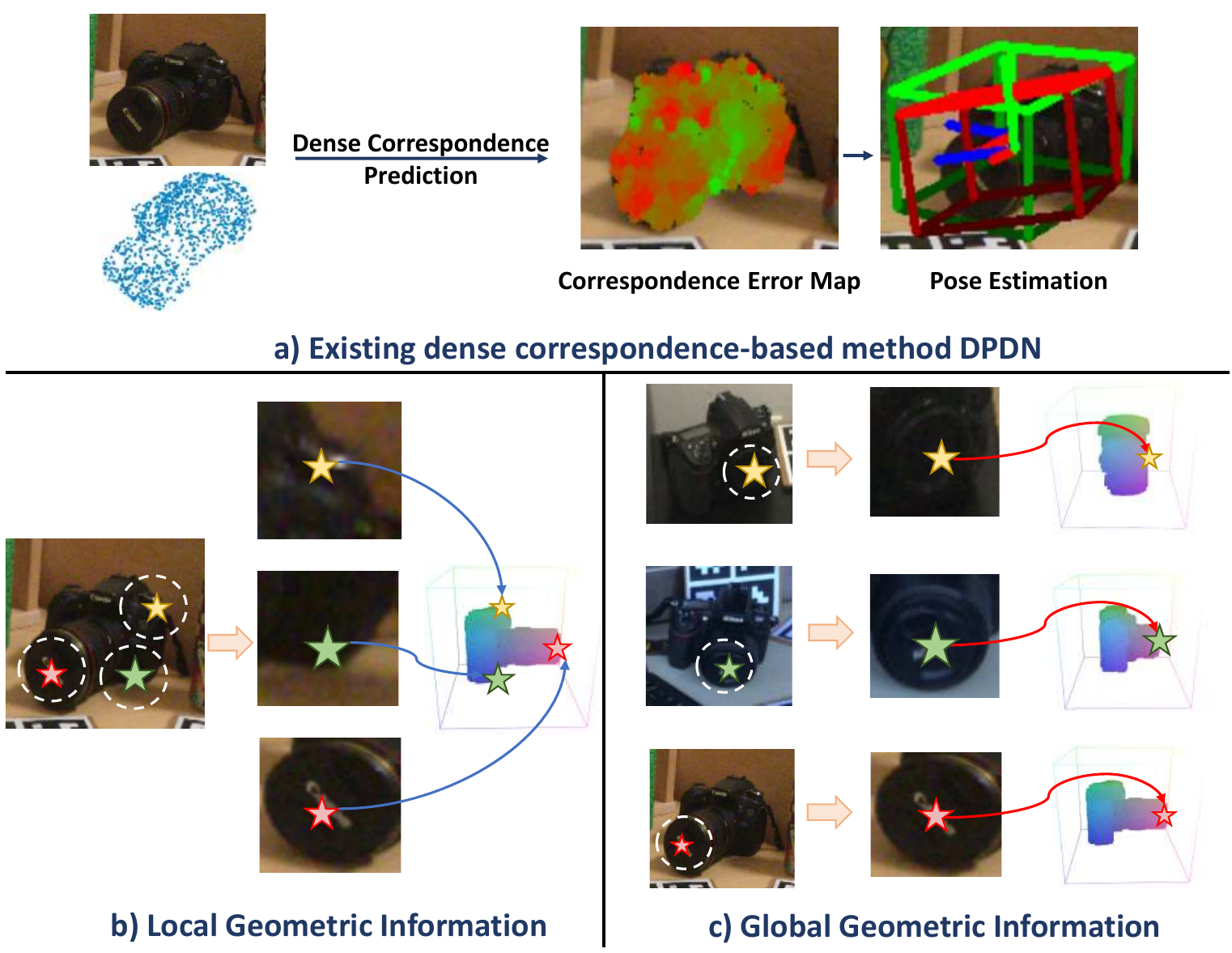}
\vspace{-0.25in}
\caption{
a) The visualization for the correspondence error map and final pose estimation 
of the dense correspondence-based method, DPDN ~\cite{lin2022dpdn}. 
Green/red indicates small/large errors and GT/predicted bounding box.
b) Points belonging to different parts of the same instance may exhibit similar visual features. 
Thus, the local geometric information is essential to distinguish them from each other. 
c) Points belonging to different instances may exhibit similar 
local geometric structures.  
Therefore, the global geometric information is crucial for correctly mapping them 
to the corresponding NOCS coordinates. 
}
\label{fig:motivation} 
\end{center}
\vspace{-0.25in}
\end{figure} 
In the category-level 6D object pose estimation task, 
networks are trained with numerous CAD models of 
different instances and required to estimate the 6D poses 
of unseen instances which belong to the same category 
but have different 3D shapes.
Consequently, the main challenge of this task lies in 
the significant intra-category shape variation. 
Most of the existing category-level methods aim to establish
dense correspondences between observed image points (RGB or RGB-D) and 
the Normalized Object Coordinate Space (NOCS) \cite{Wang2019nocs}. 
They can be categorized into two groups, i.e., 
the direct dense prediction methods and 
the two-stage deform-and-match methods.
The direct dense prediction methods train their networks to directly
map observed image points into the NOCS to 
obtain dense correspondences ~\cite{dualposenet, udacope, istnet, ttacope, Wang2019nocs}. 
On the other hand, the two-stage methods predict deformation fields 
on a categorical shape prior \cite{Tian2020spd} to reconstruct instance models first, 
and then establish dense correspondences between observed points and reconstructed models. 
With point-level dense correspondences, the final pose and size can be retrieved 
via Umeyama algorithm \cite{umeyama1991least} or regression networks. 
Although dense correspondence-based methods have made significant progress, 
they tend to generate numerous incorrect correspondences for 
instances with large shape variations (Figure~\ref{fig:motivation} (a)). 
We owe this problem to that these methods do not explicitly consider the geometric information 
of different instances, which is of significant importance for 
generalization to unseen instances. 
As shown in Figure \ref{fig:motivation} (b), 
the three points corresponding to different parts of the camera 
have similar visual features 
but distinct local structures. 
Thus, incorporating local geometric information 
is crucial to effectively distinguish them from each other. 
On the other hand, as illustrated in Figure \ref{fig:motivation} (c), 
the three points on different cameras exhibit similar local structures 
but correspond to different NOCS coordinates 
due to the large shape variations on the camera lenses. 
These points can be effectively discriminated by 
taking the global geometric information into account.
%
%

To facilitate each point with geometric information in dense correspondence-based methods, 
an intuitive way is to employ vanilla attention mechanism 
\cite{vaswani2017attention} to aggregate features from all other points. 
However, it will greatly increase the computational overhead due to the vast number of points. 
Different from existing methods, we aim to utilize a set of 
sparse keypoints to represent the shapes of different instances and extract geometric-aware keypoint features 
to establish robust keypoint-level correspondences for 6D pose estimation. Essentially, the global geometric information can be represented by the relative positions among keypoints, and the local geometric information can be represented by the relative positions between keypoints and their neighboring points. 
Nevertheless, it's non-trivial because of following challenges: 
1) The shapes of different instances vary significantly. 
Thus, the model needs to adaptively detect representive keypoints for 
different instances to comprehensively represent their shapes. 
2) The local and global geometric information are indispensable. 
How to efficiently encode such geometric information into keypoint features
needs to be carefully considered to achieve better generalization on unseen instances.

Motivated by the above discussions, we propose a novel Instance-\textbf{A}daptive and 
\textbf{G}eometric-Aware Keypoint Learning method for category-level 6D object pose estimation (AG-Pose). 
The framework is shown in Figure \ref{fig:model} (a).  
The proposed method has two key designs: 
the Instance-Adaptive Keypoint Detection (IAKD) module and 
the Geometric-Aware Feature Aggregation (GAFA) module. 
The IAKD is designed to adaptively detect keypoints for 
various instances. 
Specifically, we initialize a set of category-shared learnable queries 
as keypoint detectors. We first convert them to instance-adaptive 
detectors by aggregating the object features into learnable 
queries via attention mechanism \cite{vaswani2017attention}. 
Subsequently, the heatmap of keypoints is obtained by 
calculating similarities between instance-adaptive detectors 
and object features. 
However, the IAKD alone can not guarantee the detected keypoints to 
focus on different local parts to fully represent the shapes of different instances. 
Thus, we further design a diversity loss and an object-aware chamfer distance loss to 
constrain the distribution of keypoints, yielding dispersed and object-focused keypoints.
The GAFA module is designed to efficiently extract the local and global 
geometric information for detected keypoints. 
%
In particular, to incorporate local geometric information, 
GAFA selects the spatially nearest K points for each keypoint 
and incorporate their relative positions to aggregate features 
from its neighbors. 
To incorporate global geometric information, we further integrate global feature and relative positional embeddings into keypoint features. 
%
%
By combining these two modules together, the proposed method can efficiently learn instance-adaptive and geometric-aware keypoints to establish robust keypoint-level correspondences for pose estimation.
%

In summary, our contributions are as follows:
\begin{itemize}
    \item We propose a novel instance-adaptive and geometric-aware keypoint learning method
    for category-level 6D object pose estimation, which can better generalize to unseen instances with large shape variations. 
    To the best of our knowledge, 
    this is the first adaptive keypoint-based method 
    for category-level 6D object pose estimation.
    %
    \item We evaluate our framework on widely adopted CAMERA25 and REAL275 datasets, and results demonstrate that the proposed method sets a new state-of-the-art performance without using categorical shape priors. 
\end{itemize}

\section{Related Works} \label{sec: related_works}
\subsection{Instance-level 6D object pose estimation}
Instance-level 6D object pose estimation aims to 
estimate the pose of a known object given its CAD model. 
In recent years, numerous methods based on 
direct regression ~\cite{xiang2018posecnn, wang2019densefusion, Di2021sopose, mo2022es6d}, 
fixed keypoint detection ~\cite{ peng2019pvnet, wu2022vote, mv6d, wu2022keypointrefine}, 
and dense 2D-3D correspondences ~\cite{xu2022rnnpose, Park2019pix2pose} have been proposed. 
%
%
Direct regression-based methods take the observation image (RGB or RGBD) as input 
and directly predict the 3D rotation and translation via regression networks. 
The advantage of these methods lies in end-to-end pose estimation without extra post-processing. 
On the other hand, methods based on fixed keypoint detection 
usually predefine a set of keypoints on the object CAD model 
and detect their corresponding positions in the input image. 
Subsequently, pose estimation is achieved through 
Perspectiven-Point (PnP) algorithms. 
For example, ~\cite{peng2019pvnet, He2020pvn3d, He2021fffb6d,wu2022vote} 
utilize a dense voting strategy to detect keypoints and 
achieve state-of-the-art performance. 
Dense 2D-3D correspondence-based methods aim to predict the 
corresponding position on the object CAD model for 
every point ~\cite{Wang2021GDRN, liu2022gdrnpp_bop}, 
which are more robust to occlusion.  
Among them, the approaches based on fixed keypoints are most related to our method.
However, they adopt fixed keypoints on a given CAD model 
while our method aims to adaptively detect keypoints for different 
instances because the CAD models are not accessible during inference 
in category-level task.    
%

\subsection{Category-level 6D object pose estimation}
To improve the generalization ablity on unseen instances, 
the category-level 6D object pose estimation is introduced, 
which aims to estimate the poses of all instances 
belonging a specific category without using their CAD models. 
NOCS~\cite{Wang2019nocs} proposes to use a shared canonical representation 
called Normalized Object Coordinate Space (NOCS) to represent 
the shapes of all instances. 
They first predict the NOCS coordinates of observed points 
and then apply Umeyama~\cite{umeyama1991least} algorithm to 
recover the pose and size. 
%
%
To handle the intra-category shape variation, 
SPD~\cite{Tian2020spd} proposes a deformation and matching strategy. 
They first reconstruct instance models by deforming a 
categorical shape prior and then match observations to the reconstructed models.
%
Inspired by SPD, multiple works~\cite{chen2021sgpa, lin2022dpdn, wang2021care,zhang2022rbp} 
have been proposed to improve the processes of shape prior deformation, 
correspondence matching and et al., continuously improving 
the pose estimation performance. 
However, the above methods have not explicitly taken the local and 
global geometric information of different instances into consideration, 
which results in poor generalization ability to 
unseen instances with significant shape variations.

%
%

\section{Methodology}
\subsection{Overview}
Given a RGB-D image, following previous works \cite{Zheng2023hspose, lin2022dpdn}, 
we first employ an off-the-shelf MaskRCNN \cite{he2017maskrcnn} to 
obtain the segmentation mask and category label for each object. 
For each segmented object, we use the segmentation mask to get the cropped RGB image $\mathbf{I}_{obj} \in \mathbb{R}^{H \times W \times 3}$ and the point cloud $\mathbf{P}_{obj} \in \mathbb{R}^{N \times 3}$, 
where $N$ is the number of points and $\mathbf{P}_{obj}$ is acquired by back-projecting the cropped depth image 
using camera intrinsics followed by a downsampling process. 
By taking $\mathbf{I}_{obj}$ and $\mathbf{P}_{obj}$ as input, the proposed method 
aims to estimate the 3D rotation $\mathbf{R} \in SO(3)$, the 3D translation $\bm{t} \in \mathbb{R}^3$ 
and the size $\bm{s} \in \mathbb{R}^3$ of the target object.

\begin{figure*}
\begin{center}
\vspace{-5pt}
\includegraphics[width=1\linewidth,height=0.64375\linewidth]{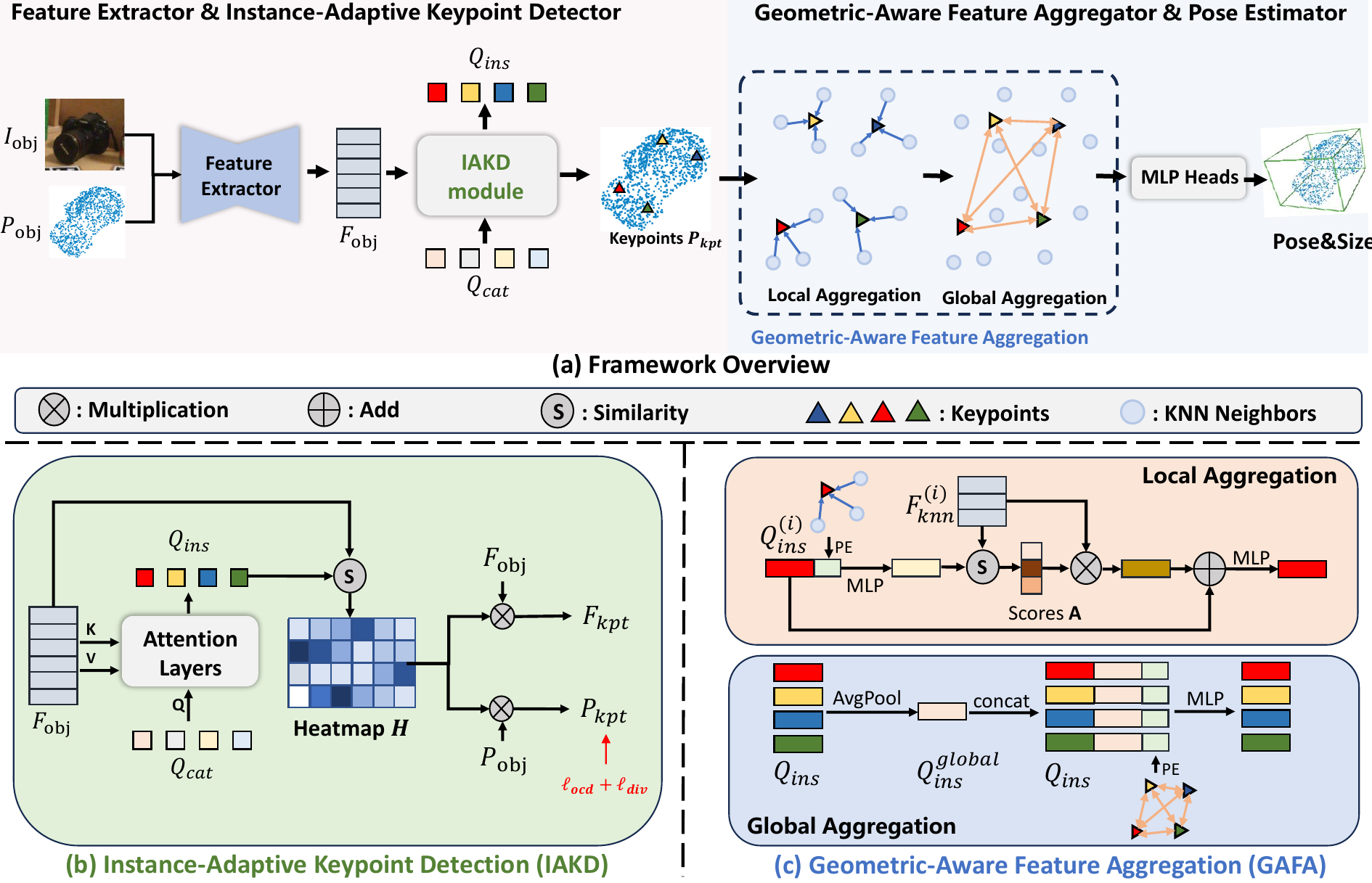}
\vspace{-0.1in}
\caption{
a) Overview of the proposed AG-Pose. 
b) Illustration of the IAKD module. We initialize a set of category-shared 
learnable queries and convert them into instance-adaptive detectors by integrating the object features. 
The instance-adaptive detectors are then used to detect keypoints for the object.  
To guide the learning of the IAKD module, we futher design the $L_{div}$ and $L_{ocd}$ to constrain the distribution of keypoints.
c) Illustration of the GAFA module. 
Our GAFA can efficiently integrate the geometric information into keypoint features through a two-stage feature aggregation process. 
}
\vspace{-0.25in}
\label{fig:model}
\end{center}
\end{figure*}

The framework of proposed AG-Pose is shown in Figure \ref{fig:model} (a).  
Our method consists of four main components: 
Feature Extractor (Sec. \ref{sec:FeatureExtractor}), 
Instance-Adaptive Keypoint Detector (Sec. \ref{sec:SparseKeypointsDetector}), 
Geometric-Aware Feature Aggregator (Sec. \ref{sec:LocalFeatureAggregator}) 
and Pose\&Size Estimator (Sec. \ref{sec:NOCSandPose&SizePredictor}). 
%
%
Details about each component are as follows.
\subsection{Feature Extractor}\label{sec:FeatureExtractor}
For the input point cloud $\mathbf{P}_{obj}$, 
we utilize the PointNet++ \cite{qi2017pointnet++} to extract 
point features $\mathbf{F}_{P} \in \mathbb{R}^{N \times C_{1}}$. 
For the RGB image $\mathbf{I}_{obj}$, following \cite{wang2019densefusion}, 
a PSP network \cite{zhao2017psp} with ResNet-18 \cite{he2016resnet} is applied 
to extract pixel-wise appearance features from $\mathbf{I}_{obj}$. 
We then choose those pixel features corresponding to $\mathbf{P}_{obj}$ to 
obtain the RGB features $\mathbf{F}_{I} \in \mathbb{R}^{N \times C_{2}}$. 
Lastly, we concatenate $\mathbf{F}_{I}$ and $\mathbf{F}_{P}$ 
to form $\mathbf{F}_{obj} \in \mathbb{R}^{N \times C}$ as the 
input for the subsequent networks.

\subsection{Instance-Adaptive Keypoint Detector}\label{sec:SparseKeypointsDetector}
%
%
As introduced in Section~\ref{sec:intro},
the geometric information is indispensable to establish 
robust correspondences.
An intuitive idea to facilitate each point with geometric information is to employ the vanilla attention mechanism \cite{vaswani2017attention}
to aggregate features from all other points. 
But it is computationally expensive due to the 
large number of points.  
Instead, we aim to utilize a set of sparse keypoints to 
represent the shapes of different instances for pose estimation.
However, since the shapes vary across different instances and the instance models are not accessible during inference, 
fixed keypoint detection methods like ~\cite{He2020pvn3d,He2021fffb6d,peng2019pvnet,wu2022vote} 
are not applicable.
Besides, methods like Farest Point Sampling are not end-to-end trainable, 
which may detect keypoints on outlier points caused by the 
noisy segmentation mask, the incorrect depth values and et al.    
Consequently, we design an Instance-Adaptive Keypoint Detection 
module to adaptively detect keypoints for 
instances with different shapes which can efficiently represent the object and avoid 
focusing on outlier points. 

%

\begin{figure}[t]
    \begin{center}
    \vspace{-5pt}
    \includegraphics[width=0.85\linewidth]{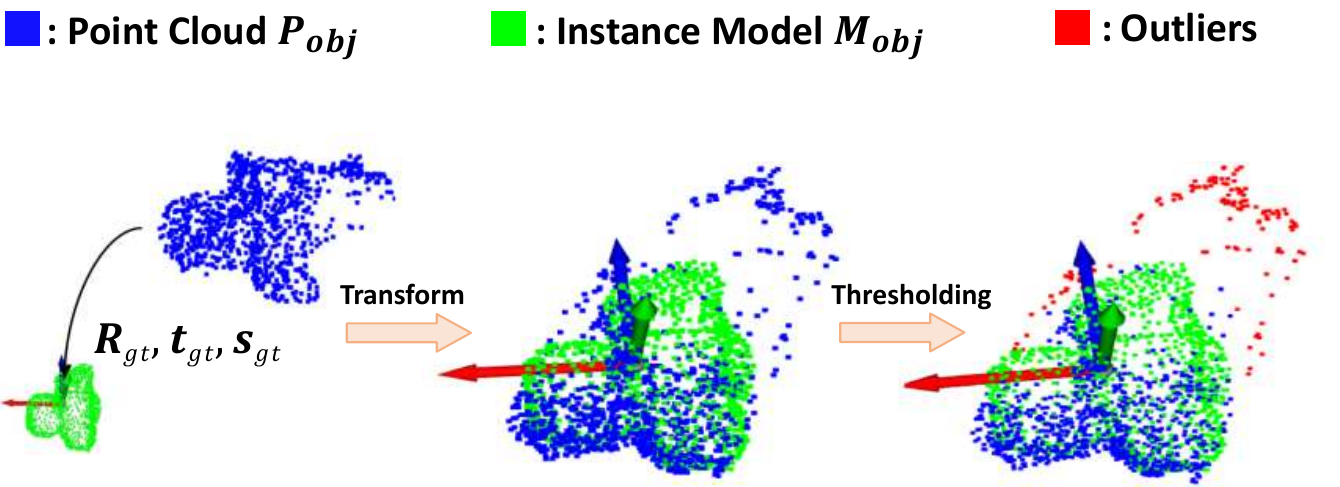}
    \vspace{-0.075in}
    \caption{Illustration of the outlier filter process.}
    \vspace{-0.25in}
    \label{fig:filter}
    \end{center}
\end{figure}

The pipeline of IAKD is illustrated in Figure \ref{fig:model} (b). 
Specifically, we initialize a set of 
category-shared learnable queries 
$\mathbf{Q}_{cat} \in \mathbb{R}^{N_{kpt} \times C}$ 
and each of them represents a keypoint detector. 
We first inject the object features $\mathbf{F}_{obj}$ 
into learnable queries via \textit{Transformer Layers} \cite{vaswani2017attention}. 
This process converts the category-shared detectors $\mathbf{Q}_{cat}$ into 
instance-adaptive detectors $\mathbf{Q}_{ins} \in \mathbb{R}^{N_{kpt} \times C}$ 
conditioned on object features $\mathbf{F}_{obj}$. In detail, 
\begin{gather}
\mathbf{Q} =  \mathbf{Q}_{cat}\mathbf{W}^{q}, \mathbf{K} =  \mathbf{F}_{obj}\mathbf{W}^{k}, \mathbf{V} =  \mathbf{F}_{obj}\mathbf{W}^{v},
\\
\mathbf{Q}_{ins} =\textit{Norm} (\mathbf{Q}_{cat} 
 + \textit{softmax} (\mathbf{Q} \mathbf{K}^{T} ) \mathbf{V}). 
\end{gather}
Then we calculate the cosine similarities between the 
instance-adaptive detectors 
$\mathbf{Q}_{ins}$ and the 
object features $\mathbf{F}_{obj}$ to generate keypoint heatmap $\mathbf{H} \in \mathbb{R}^{N_{kpt} \times N}$. 
%
%
Subsequently, the 3D coordinates of keypoints $\mathbf{P}_{kpt} \in \mathbb{R}^{N_{kpt} \times 3}$ 
and their corresponding features $\mathbf{F}_{kpt} \in \mathbb{R}^{N_{kpt} \times C}$ 
under camera space are obtained by weighted sum as:
\begin{gather}
\mathbf{P}_{kpt} = \textit{softmax}(\mathbf{H})\times \mathbf{P}_{obj}, \\
\mathbf{F}_{kpt} = \textit{softmax}(\mathbf{H})\times \mathbf{F}_{obj}.
\end{gather}
%
Since our target is to utilize a set of sparse keypoints 
to represent the geometric information of the object,
the detected keypoints should be well distributed 
on the surface of the object.
However, we find that the keypoints tend to cluster in small regions and often focus on non-surface or outlier points without explicit supervision.
To encourage the keypoints to focus on different parts, we futher 
design a diversity loss $L_{div}$ to 
force the detected keypoints to be away from each other. In detail, 
\begin{gather}
    L_{div} = \sum_{i=1}^{N_{kpt}} \sum_{j=1, j \neq i}^{N_{kpt}} \mathbf{d}(\mathbf{P}_{kpt}^{(i)},\mathbf{P}_{kpt}^{(j)}),
    \\
    \mathbf{d}(\mathbf{P}_{kpt}^{(i)},\mathbf{P}_{kpt}^{(j)}) = 
     \max \{th_{1} - \|\mathbf{P}_{kpt}^{(i)}-\mathbf{P}_{kpt}^{(j)}\|_{2}, 0 \},
    \end{gather}
where $th_{1}$ is a hyperparameter and 
$\mathbf{P}_{kpt}^{(i)}$ stands for the $i$-th keypoint. 
To encourage the keypoints to locate on the surface of the object and exclude outliers simultaneously, 
we design an object-aware chamfer distance loss $L_{ocd}$ to constrain the distribution of $\mathbf{P}_{kpt}$. Formally, we first utilize the ground truth pose and size 
$\mathbf{R}_{gt},\mathbf{t}_{gt},\mathbf{s}_{gt}$ to transform $\mathbf{P}_{obj}$ to the NOCS and remove outlier points based on instance model $\mathbf{M}_{obj} \in \mathbb{R}^{M \times 3}$ to produce $\mathbf{P}_{obj}^{\star}$. In formula, \begin{gather}
\mathbf{P}_{obj}^{\star} = \{x_i | x_{i} \in \mathbf{P}_{obj} \quad \textit{and} 
\nonumber\\ 
\min_{y_{j} \in \mathbf{M}_{obj}} \|\frac{1}{\|\mathbf{s}_{gt}\|_{2}}\mathbf{R}_{gt}(x_{i} - \mathbf{t}_{gt}) - y_{j}\|_{2} < th_{2} \}. 
\end{gather}
The outlier filter process is also illustrated in Figure \ref{fig:filter}. Then the $L_{ocd}$ is calculated as follows,  
\begin{gather}
L_{ocd} = \frac{1}{|\mathbf{P}_{kpt}|} \sum_{x_i \in \mathbf{P}_{kpt}} \min_{y_{j} \in \mathbf{P}_{obj}^{\star}} \|x_i - y_j \|_{2}. 
\end{gather}
By constraining keypoints to be close to $\mathbf{P}_{obj}^{\star}$, 
the IAKD module can automatically learn to filter out outlier points during inference.
\subsection{Geometric-Aware Feature Aggregator}\label{sec:LocalFeatureAggregator}
With the detected keypoints for each object,
a straightforward way is to predict the NOCS coordinates for these keypoints to establish keypoint-level correspondences.
However, the keypoint features are lack of geometric information,
which can result in numerous incorrect correspondences on unseen instances. 
%
Thus, we propose a Geometric-Aware Feature Aggregation module to efficiently incorporate geometric information into keypoint features. 

The GAFA adopts a two-stage pipeline for geometric feature aggregation, 
as illustrated in Figure \ref{fig:model} (c). 
We first select the nearest $K$ neighbors in $\mathbf{P}_{obj}$ 
and their corresponding features in $\mathbf{F}_{obj}$ for each keypoint 
to obtain $\mathbf{P}_{knn} \in \mathbb{R}^{N_{kpt} \times K \times 3}$ 
and $\mathbf{F}_{knn} \in \mathbb{R}^{N_{kpt} \times K \times C}$. 
Essentially, the global geometric information can be represented by the relative positions among keypoints, and the local geometric information can be represented by the relative positions between keypoints and their neighboring points. 
Thus, we utilize the relative positional embeddings $\alpha$ and $\beta$ between 
keypoints and their neighbors to represent the 
local geometric features $f_{l}$ and global geometric features $f_{g}$, respectively. In detail, 
\begin{gather}
    \alpha_{i,j} = MLP(\mathbf{P}_{kpt}^{(i)} - \mathbf{P}_{knn}^{(i,j)}),
    f_{l}^{(i)} = \textit{AvgPool}(\alpha_{i,:}), 
    \\
    \beta_{i,j} = MLP(\mathbf{P}_{kpt}^{(i)} - \mathbf{P}_{kpt}^{(j)}), 
    f_{g}^{(i)} = \textit{AvgPool}(\beta_{i,:}),
\end{gather}
where $f_{l}^{(i)},f_{g}^{(i)} \in \mathbb{R}^{1 \times C}$ and $\mathbf{P}_{knn}^{(i,j)}$ 
is the $j$-th neighboring point of $\mathbf{P}_{kpt}^{(i)}$. To incorporate local geometric information into keypoints,    
we combine the keypoint features $\mathbf{Q}_{ins}$ with $f_{l}$ and 
calculate local correlation scores $\mathbf{A}$ between keypoints and neighboring points, 
which is then used to aggregate features from neighbors. 
The local feature aggregation process for $i$-th keypoint feature $\mathbf{Q}_{ins}^{(i)}$ is as follows,
\begin{gather}
\mathbf{A} = sim(MLP(\textit{cat}\left[ \mathbf{Q}_{ins}^{(i)}, f_{l}^{(i)}\right]), \mathbf{F}_{knn}^{(i)}), 
\\
\mathbf{Q}_{ins}^{(i)} = MLP(softmax(\mathbf{A}) \times \mathbf{F}_{knn}^{(i)} + \mathbf{Q}_{ins}^{(i)}). 
\end{gather}
The above process is executed for all keypoints in parallel to extract representive
local geometric features. 
However, as shown in Figure \ref{fig:motivation} (c), the local geometric information alone 
is insufficient to handle significant shape variation across different instances. 
Therefore, we futher inject the global geometric feature 
into $\mathbf{Q}_{ins}$ by leveraging all keypoint 
features and the global geometric features $f_{g}$. 
In formula, 
\begin{gather}
\mathbf{Q}_{ins}^{global} = \textit{AvgPool}(\mathbf{Q}_{ins}),
\\
\mathbf{Q}_{ins}^{(i)} = MLP(\textit{concat}\left[ \mathbf{Q}_{ins}^{(i)}, \mathbf{Q}_{ins}^{global}, f_{g}^{(i)} \right]),
\end{gather}
where $\mathbf{Q}_{ins}^{global} \in \mathbb{R}^{1 \times C}$ 
is the global feature of $\mathbf{Q}_{ins}$. 

The above two-stage aggregation allows keypoints to adaptively 
aggregate local geometric features from neighbors and 
global geometric information from 
other keypoints. 
It's worth noting that since the number of keypoints is small, 
the above process is computationally efficient.

\subsection{Pose\&Size Estimator}\label{sec:NOCSandPose&SizePredictor}
After obtaining geometric-aware keypoint features, 
following previous work \cite{lin2022dpdn}, we use MLP to predict the NOCS coordinates of keypoints 
$\mathbf{P}_{kpt}^{nocs} \in \mathbb{R}^{N_{kpt} \times 3}$ 
from $\mathbf{Q}_{ins}$ and regress the final pose and size 
$\mathbf{R}, \mathbf{t}, \mathbf{s}$ via 
this set of keypoint-level correspondences. In formula,  
\begin{gather}
\mathbf{P}_{kpt}^{nocs} = MLP(\mathbf{Q}_{ins}),
\\
\mathbf{f}_{pose} = \textit{concat}\left[ \mathbf{P}_{kpt},\mathbf{F}_{kpt},\mathbf{P}_{kpt}^{nocs}, \mathbf{Q}_{ins} \right],
\\
\mathbf{R}, \mathbf{t}, \mathbf{s} = MLP_{R}(\mathbf{f}_{pose}), MLP_{t}(\mathbf{f}_{pose}),
MLP_{s}(\mathbf{f}_{pose}).
\end{gather}
For more details please refer to \cite{lin2022dpdn}. 
We use the 6D representation \cite{zhou2019continuity6d} for the rotation $\mathbf{R}$. 
And for the translation $\mathbf{t}$ we follow \cite{Zheng2023hspose} to predict the residual translation between the ground truth and the mean value of the point cloud.

\subsection{Overall Loss Function}\label{sec:Overall Loss Function}
The overall loss function is as follows:
\begin{equation}
L_{all} = \lambda_{1} L_{ocd} + \lambda_{2} L_{div} + \lambda_{3} L_{nocs} + \lambda_{4} L_{pose},
\end{equation}
where $\lambda_{1}, \lambda_{2}, \lambda_{3}, \lambda_{4}$ are 
hyperparameters to balance the contribution of each term. 
For $L_{pose}$, we simply use $L_{1}$ loss as follows, 
\begin{equation}
L_{pose} = \|\mathbf{R}_{gt} - \mathbf{R}\|_{2} + \|\mathbf{t}_{gt} - \mathbf{t}\|_{2} + \|\mathbf{s}_{gt} - \mathbf{s}\|_{2}.
\end{equation}
We generate ground truth NOCS coordinates of 
keypoints $\mathbf{P}_{kpt}^{gt}$ by projecting their 
coordinates under camera space $\mathbf{P}_{kpt}$ into NOCS using the ground-truth 
$\mathbf{R}_{gt}, \mathbf{t}_{gt}, \mathbf{s}_{gt}$. 
For $L_{nocs}$, we use the $SmoothL_{1}$ loss \cite{lin2022dpdn}. In formula, 
\begin{gather}
\mathbf{P}_{kpt}^{gt} = \frac{1}{\|\mathbf{s}_{gt}\|_{2}}\mathbf{R}_{gt}(\mathbf{P}_{kpt} - \mathbf{t}_{gt}),
\\
L_{nocs} = SmoothL_{1} (\mathbf{P}_{kpt}^{gt},\mathbf{P}_{kpt}^{nocs}).
\end{gather}

\begin{table*}
\caption{\textbf{
    Quantitative comparisons with state-of-the-art methods on the REAL275 dataset.}
    }
\label{tab:REAL275}
\centering
\begin{tabular}{l|c|cc|cccc}
\hline
Method    & Use of Shape Priors & $IoU_{50}$         & $IOU_{75}$         & $5^{\circ} \, 2\, \textbf{cm}$         & $5^{\circ} \, 5 \, \textbf{cm}$         & $10^{\circ} \, 2 \, \textbf{cm}$        & $10^{\circ} \, 5 \, \textbf{cm}$        \\ \hline
NOCS \cite{Wang2019nocs}     & \ding{55}                  & 78            & 30.1          & 7.2           & 10            & 13.8          & 25.2          \\
DualPoseNet \cite{dualposenet} & \ding{55}                   & 79.8          & 62.2          & 29.3          & 35.9          & 50            & 66.8          \\
GPV-Pose \cite{di2022gpv}  & \ding{55}                   & ---          & 64.4          & 32            & 42.9          & ---          & 73.3          \\
IST-Net  \cite{istnet}  & \ding{55}                   & 82.5          & 76.6          & 47.5          & 53.4          & 72.1          & 80.5          \\ 
Query6DoF \cite{query6dof}    & \ding{55}                  & 82.5          & 76.1          & 49            & 58.9          & 68.7          & 83            \\\hline
SPD  \cite{Tian2020spd}     & \ding{51}                  & 77.3          & 53.2          & 19.3          & 21.4          & 43.2          & 54.1          \\
SGPA \cite{chen2021sgpa}     & \ding{51}                  & 80.1          & 61.9          & 35.9          & 39.6          & 61.3          & 70.7          \\
SAR-Net \cite{lin2022sar}   & \ding{51}                  & 79.3          & 62.4          & 31.6          & 42.3          & 50.3          & 68.3          \\
RBP-Pose \cite{zhang2022rbp}   & \ding{51}                  & ---          & 67.8          & 38.2          & 48.1          & 63.1          & 79.2          \\
DPDN \cite{lin2022dpdn}     & \ding{51}                  & 83.4          & 76            & 46            & 50.7          & 70.4          & 78.4          \\ \hline
AG-Pose      & \ding{55}                  & \textbf{83.7} & \textbf{79.5} & \textbf{54.7} & \textbf{61.7} & \textbf{74.7} & \textbf{83.1} \\ \hline
\end{tabular}
\end{table*}
\section{Experiments}
\textbf{Datasets.} 
Following previous works \cite{lin2022dpdn, query6dof,istnet}, we conduct experiments on CAMERA25 and REAL275 \cite{Wang2019nocs}, 
the most widely adopted datasets for category-level object pose estimation. 
CAMERA25 is a synthetic RGB-D dataset that contains 300K 
synthetic images of 1,085 instances from 6 different categories. 
In this dataset, 25,000 images of 184 instances are used for evaluation and the others are used for training. 
REAL275 dataset is a more challenging real-world dataset which shares the same 
categories with CAMERA25. 
It comprises 7K images from 13 different scenes. 
2,750 images of 6 scenes are left for validation, including 3 unseen instances per category.

\textbf{Implementation details.} 
For a fair comparison, we use the same segmentation masks as SPD~\cite{Tian2020spd} and DPDN~\cite{lin2022dpdn} from 
MaskRCNN \cite{he2017maskrcnn}. 
For the data preprocessing,
images are cropped and resized to $192 \times 192$ before feature extraction,
and the number of points $N$ in point cloud is set as 1024. 
For model parameters,
the number of keypoints is set as $N_{kpt}=96$,
and the local range for each keypoint in GAFA is set as $K=16$.  
The feature dimensions are set as $C_{1}=128$, $C_{2}=128$ and $C=256$, respectively. 
For the hyper-parameters in the loss functions,
the two thresholds in LQKD are set as $th_{1}=0.01$ and $th_{2}=0.1$,
and $\lambda_{1},\lambda_{2},\lambda_{3},\lambda_{4}$ are 1.0,5.0,1.0,0.3, respectively. 
For model optimizing,
we train our model on both synthetic and real datasets for evaluation on REAL275  
while only on synthetic dataset for evaluation on CAMERA25. 
Following previous work~\cite{lin2022dpdn}, we use random translation $\Delta \mathbf{t} \sim \textit{U}(-0.02, 0.02)$ and 
scaling $\Delta \mathbf{s} \sim \textit{U}(-0.8, 1.2)$ for data augmentation and apply random rotation for each axis with 
rotation degree sampled from $\textit{U}(0, 20)$. 
We train our network using the ADAM \cite{kingma2014adam} optimizer and employ the 
triangular2 cyclical learning rate schedule \cite{smith2017cyclical} 
ranging from 2e-5 to 5e-4. 
All experiments are conducted on a single 
RTX3090Ti GPU with a batch size of 24. 

\textbf{Evaluation metrics.}
Following previous works~\cite{Tian2020spd, Wang2019nocs}, 
we evaluate the model performance with two metrics.
\begin{itemize}
    \item 3D IoU. We report the mean precision of Intersection-over-Union (IoU) for 3D bounding boxes 
    with thresholds of $50\%$ and $75\%$. 
    This metric incorporates both the pose and size of the object.

    \item $n^{\circ} \, m \, \textbf{cm}$ is used for direct 
    evaluation of the rotation and translation errors. 
    Only predictions with rotation error less than $n^{\circ}$ 
    and translation error less than $m \, \textbf{cm}$ are considered correct.
\end{itemize}

\subsection{Comparison with State-of-the-Art Methods}\label{sec:Comparisons with Existing}

\textbf{Results on REAL275 dataset.} 
The comparisons between our AG-Pose and existing 
state-of-the-art methods on the challenging REAL275 dataset are shown in Table \ref{tab:REAL275}. 
As demonstrated by the results, our AG-Pose outperforms all 
previous methods in all metrics on REAL275 dataset.
It should be noted that our method does not require the use of shape priors. 
In detail, on the most rigorous metric of $5^{\circ} \, 2 \, \textbf{cm}$, 
AG-Pose achieves the precision of $54.7\%$, surpassing the current state-of-the-art shape prior-based 
method DPDN \cite{lin2022dpdn} with a large margin by $8.7\%$. 
As for prior-free methods, our method surpasses Query6DoF \cite{query6dof}, 
IST-Net \cite{istnet} and GPV-Pose \cite{di2022gpv} 
by $5.7\%$, $7.2\%$, and $22.7\%$, respectively.
It should be noted that Query6DoF aims to learn a set of sparse queries as implicit shape priors,
and the query features are used to update the point features 
to establish better dense correspondences.
Different from it, the proposed AG-Pose aims to use 
a sparse set of keypoints to explicitly model the 
geometric information of objects to establish robust keypoint-level 
correspondences for pose estimation. 
The superior performance of our method indicates the importance of geometric information in 
category-level 6D object pose estimation.
\begin{table*}
\caption{\textbf{
    Quantitative comparisons with state-of-the-art methods on the CAMERA25 dataset.}
    }
\label{tab:CAMERA25}
\centering
\begin{tabular}{l|c|cc|cccc}
\hline
Method   & Use of Shape Prior & $IoU_{50}$         & $IoU_{75}$        & $5^{\circ} \, 2 \, \textbf{cm}$        & $5^{\circ} \, 5 \, \textbf{cm}$         & $10^{\circ} \, 2 \, \textbf{cm}$      & $10^{\circ} \, 5 \, \textbf{cm}$        \\ \hline
NOCS \cite{Wang2019nocs}     & \ding{55}                  & 83.9          & 69.5          & 32.3          & 40.9          & 48.2        & 64.4          \\
DualPoseNet \cite{dualposenet} & \ding{55}                    & 92.4          & 86.4          & 64.7          & 70.7          & 77.2        & 84.7          \\
GPV-Pose \cite{di2022gpv}  & \ding{55}                   & 93.4          & 88.3          & 72.1          & 79.1          & ---       & 89            \\ 
Query6DoF \cite{query6dof}    & \ding{55}                 & 91.9          & 88.1          & \textbf{78}   & \textbf{83.1} & 83.9        & 90            \\ \hline
SPD \cite{Tian2020spd}      & \ding{51}                  & 93.2          & 83.1          & 54.3          & 59            & 73.3        & 81.5          \\
SGPA \cite{chen2021sgpa}     & \ding{51}                  & 93.2          & 88.1          & 70.7          & 74.5          & 82.7        & 88.4          \\
SAR-Net \cite{lin2022sar}   & \ding{51}                  & 86.8          & 79            & 66.7          & 70.9          & 75.3        & 80.3          \\
RBP-Pose \cite{zhang2022rbp}     & \ding{51}                  & 93.1          & 89            & 73.5          & 79.6          & 82.1        & 89.5          \\ \hline
AG-Pose     & \ding{55}                    & \textbf{93.8} & \textbf{91.3} & 77.8 & 82.8   & \textbf{85.5} & \textbf{91.6} \\ \hline
\end{tabular}
\end{table*}

\textbf{Results on CAMERA25 dataset.} 
%
Table \ref{tab:CAMERA25} shows the quantitative results of the proposed AG-Pose on CAMERA25 dataset. 
Our method achieves the best performance under most of metrics. 
In detail, the proposed AG-Pose outperforms the state-of-the-art method Query6DoF \cite{query6dof} by 
$1.9\%$ on $IoU_{50}$, $3.2\%$ on $IoU_{75}$, $1.6\%$ on $10^{\circ} \, 2 \, \textbf{cm}$ 
and $1.6\%$ on $10^{\circ} \, 5 \, \textbf{cm}$, respectively. 
Our method achieves comparable performance with Query6DoF on 
$5^{\circ} \, 2 \, \textbf{cm}$ and $5^{\circ} \, 5 \, \textbf{cm}$ (lower by $0.2\%$ and $0.3\%$, respectively).  

\begin{figure}[t]
\begin{center}
\includegraphics[width=1\linewidth,height=0.5\linewidth]{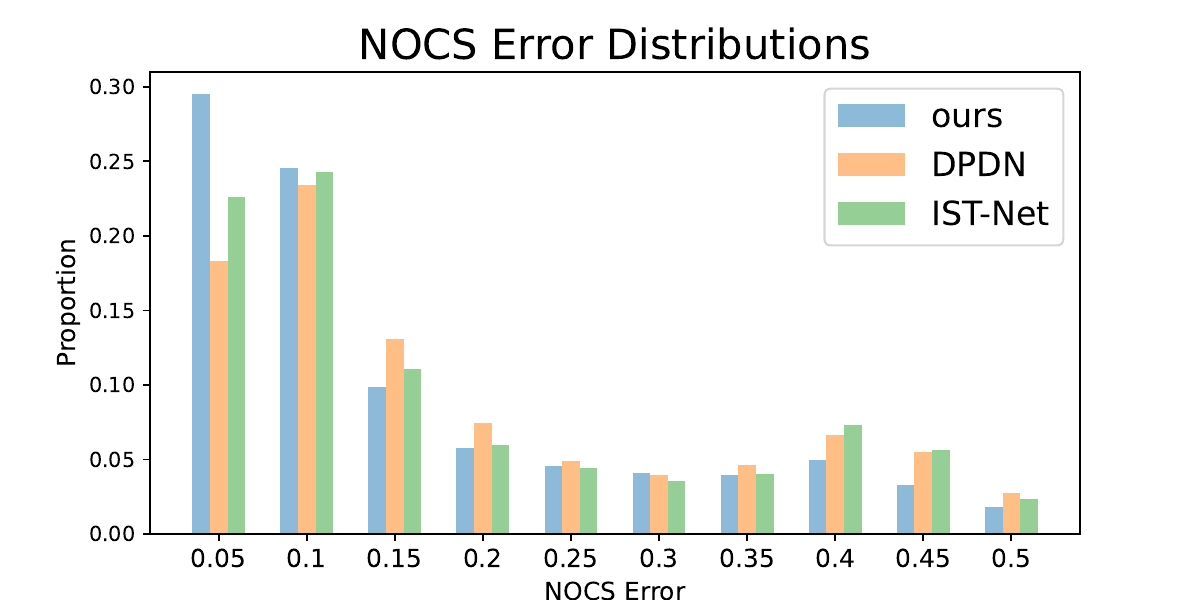}
\vspace{-0.25in}
\caption{\textbf{Comparisons of NOCS error distributions.}}
\label{fig:nocs_err} 
\end{center}
\vspace{-0.25in}
\end{figure}

\textbf{Results of correspondence errors.} 
To validate that the keypoint-level correspondences of AG-Pose are 
more accurate than point-level correspondences, 
we calculate the NOCS error distributions of 
DPDN \cite{lin2022dpdn}, ISTNet \cite{istnet} and our method 
on the REAL275 validation set. 
As shown in Figure \ref{fig:nocs_err}, 
the NOCS error of our keypoint-level correspondences is concentrated more on 
the interval from 0 to 0.1, 
while errors of dense correspondence-based methods are concentrated more on 
the interval from 0.15 to 0.5.  
It proves that the keypoint-level correspondences produced by AG-Pose 
exhibit better generalizability on unseen instances.

\subsection{Ablation Studies}\label{sec:Ablation Studies}
In this section, we conduct ablation experiments to demonstrate the 
effectiveness of the proposed method on the REAL275 dataset. 
\begin{table}
    \caption{\textbf{
        Comparisons between the IAKD and FPS.}
        }
    \label{tab:ab FPS}
    \centering
    \begin{tabular}{c|cccc}
    \hline
    Setting & $5^{\circ} \, 2 \, \textbf{cm}$         & $5^{\circ} \, 5 \, \textbf{cm}$          & $10^{\circ} \, 2 \, \textbf{cm}$         & $10^{\circ} \, 5 \, \textbf{cm}$         \\ \hline
    FPS     & 46.2            & 55.5          & 67.0          & 80.2            \\
    IAKD     & \textbf{54.7} & \textbf{61.7} & \textbf{74.7} & \textbf{83.1} \\\hline
    \end{tabular}
    \end{table}
\begin{table}
\caption{\textbf{
    Ablation studies on the number of keypoints.}
    }
\label{tab:ab kptnum}
\centering
\begin{tabular}{c|cccc}
\hline
$N_{kpt}$ & $5^{\circ} \, 2 \, \textbf{cm}$         & $5^{\circ} \, 5 \, \textbf{cm}$          & $10^{\circ} \, 2 \, \textbf{cm}$         & $10^{\circ} \, 5 \, \textbf{cm}$         \\ \hline
16     & 47.9          & 55.1          & 68.8          & 79.8          \\
32     & 48.8          & 55.7          & 73.1          & 82.9          \\
64     & 51            & 57.2          & 72.8          & 82            \\
96     & \textbf{54.7} & \textbf{61.7} & \textbf{74.7} & 83.1          \\
128    & 52.8          & 59.9          & 74.3          & \textbf{83.7} \\ \hline
\end{tabular}
\end{table}
\begin{table}
\caption{\textbf{
    Ablation studies on the proposed loss functions.}
    }
\label{tab:ab loss}
\centering
\begin{tabular}{c|cccc}
\hline
Loss     & $5^{\circ} \, 2 \, \textbf{cm}$         & $5^{\circ} \, 5 \, \textbf{cm}$         & $10^{\circ} \, 2 \, \textbf{cm}$        & $10^{\circ} \, 5 \, \textbf{cm}$        \\ \hline
$L_{div}$+$L_{ocd}$ & \textbf{54.7} & \textbf{61.7} & \textbf{74.7} & \textbf{83.1} \\
$L_{div}$+$L_{ucd}$  & 49.8          & 57.3          & 74.4          & 82.0          \\
$L_{div}$      & 46.4          & 53            & 71            & 81.3          \\
$L_{ocd}$     & 30         & 36.1          & 55.0          & 68.6          \\
None           & 29.3          & 35.6          & 56.4          & 69.6          \\ \hline
\end{tabular}
\end{table}

\textbf{Effects of the IAKD module.} 
%
To evaluate the effectiveness of the proposed Instance-Adaptive Keypoint Detection module, 
we first replace the IAKD with the most widely adopted Farest Point Sampling (FPS) strategy.
Specifically,
we use FPS to sample the same number of keypoints and use their 
corresponding features in $\mathbf{F}_{obj}$ as $\mathbf{Q}_{ins}$.
For a fair comparison, we add extra MLPs to ensure 
that the number of parameters remain unchanged. 
As shown in Table \ref{tab:ab FPS}, 
the proposed IAKD outperforms FPS on all metrics. 
We owe the advantage to the end-to-end trainable property of IAKD 
and the optimized distribution of keypoints brought by the proposed losses. 

\textbf{Effects of different keypoint numbers.} 
In Table \ref{tab:ab kptnum}, we show the impact of different keypoint numbers $N_{kpt}$. 
Surprisingly, the result shows that our method can achieve comparable 
performance ($47.9\%$ on $5^{\circ} \, 2 \, \textbf{cm}$) with the SOTA methods 
by using only a very small number ($N_{kpt}=16$) of keypoints, which demonstrates 
the superiority of proposed AG-Pose.
And as the number of keypoints ($N_{kpt}$) increases, 
the performance of our method continues to improve. 
We choose $N_{kpt}=96$ in our method for the balance between efficiency and accuracy.
%

\textbf{Effects of the $L_{div}$ and the $L_{ocd}$.} 
In our method, we utilize the proposed $L_{div}$ and $L_{ocd}$ 
to encourage the keypoints to be well distributed on the surface of the object. 
To verify the effectiveess of them, we conduct ablation experiments on these losses and the results 
are shown in Table \ref{tab:ab loss}. 
%
%
By replacing the proposed $L_{ocd}$ with the normal chamfer distance loss $L_{ucd}$, 
the performance of our model drops by $4.5\%$ on $5^{\circ} \, 2 \, \textbf{cm}$. 
The reason is that our object-aware chamfer distance loss 
can prevent the model from being affected by outliers, 
thereby improving the performance. 
The accuracy of the model further declines by $8.3\%$ 
after we remove the $L_{ocd}$. It is because $L_{ocd}$ encourages keypoints to distribute 
on the surface of objects, which can better represent the shapes of objects.  
Without $L_{div}$, the model performance declines significantly by $24.7\%$, since $L_{div}$ is indispensable for 
keypoints to focus on distinct regions of objects. 
The excessive clustering of keypoints can result in a degradation in model performance.
\begin{table}
\caption{\textbf{
    Ablation study on two-stage feature aggregation.}
    }
    \vspace{-5pt}
\label{tab:ab_two_stage}
\centering
\begin{tabular}{c|cccc}
\hline
Setting  & $5^{\circ} \, 2 \, \textbf{cm}$         & $5^{\circ} \, 5 \, \textbf{cm}$       & $10{\circ} \, 2 \, \textbf{cm}$        & $10^{\circ} \, 5 \, \textbf{cm}$        \\ \hline
Full     & \textbf{54.7} & \textbf{61.7} & \textbf{74.7} & \textbf{83.1}          \\\hline
w/o GAFA & 47.1          & 55.3        & 70.2          & 80.9          \\
w/o Local & 49          & 57.8        & 71.2         & 82.2          \\
w/o Global & 50.1          & 55.8        & 74.5          & 82.7 \\
w/ vanilla attn & 53          & 61        & 72.2          & 82.1 \\\hline
\end{tabular}
\end{table}
\begin{table}
\caption{\textbf{
    Ablation study on proposed GAFA.}
    }
    \vspace{-5pt}
\label{tab:ab LGFA}

\centering
\begin{tabular}{c|cccc}
\hline
         & $5^{\circ} \, 2 \, \textbf{cm}$         & $5^{\circ} \, 5 \, \textbf{cm}$       & $10{\circ} \, 2 \, \textbf{cm}$        & $10^{\circ} \, 5 \, \textbf{cm}$        \\ \hline
K=8      & 49.9          & 57.6        & 73.1          & 82.5          \\
K=16     & \textbf{54.7} & \textbf{61.7} & \textbf{74.7} & 83.1          \\
K=24     & 54.1          & 61.1        & 73.7          & \textbf{83.2} \\
K=32     & 52.7          & 59.9        & 73.6          & 82.8          \\ \hline
\end{tabular}
\vspace{-15pt}
\end{table}

\textbf{Effects of the GAFA module.}
Here we conduct ablation studies on the Geometric-Aware Feature Aggregation module. 
For a fair comparison, we replace the GAFA with MLPs that have the same number of parameters. 
The results are shown in Table \ref{tab:ab_two_stage}. 
In particular, we observe that performance drops by $7.6\%, 4.6\%$ and $5.7\%$ 
on $5^{\circ} \, 2 \, \textbf{cm}$
when removing the whole GAFA, the global feature aggregation 
and the local feature aggregation, respectively.
As discussed in section \ref{sec:intro}, both local and global geometric information 
play crucial roles in establishing accurate correspondences on unseen objects. 
Our GAFA can effectively encode such geometric information into keypoint features, 
thus achieving better accuracy. 
Futhermore, we replace the geometric-aware feature aggregation operation with vanilla attention mechanism (w/ vanilla attn) in the GAFA. 
The experimental result shows that the proposed geometric-aware feature aggregation achieve better performance 
because the geometric information is indispensable in category-level 6D object pose estimation.
Last, we explore the influence of the local aggregation range $K$ 
on the accuracy of our model in Table \ref{tab:ab LGFA}.  
The results demonstrate that $K=16$ yields the most significant performance gains. 

\begin{figure}
    \begin{center}
    \vspace{-5pt}
    \includegraphics[width=1\linewidth,height=0.47\linewidth]{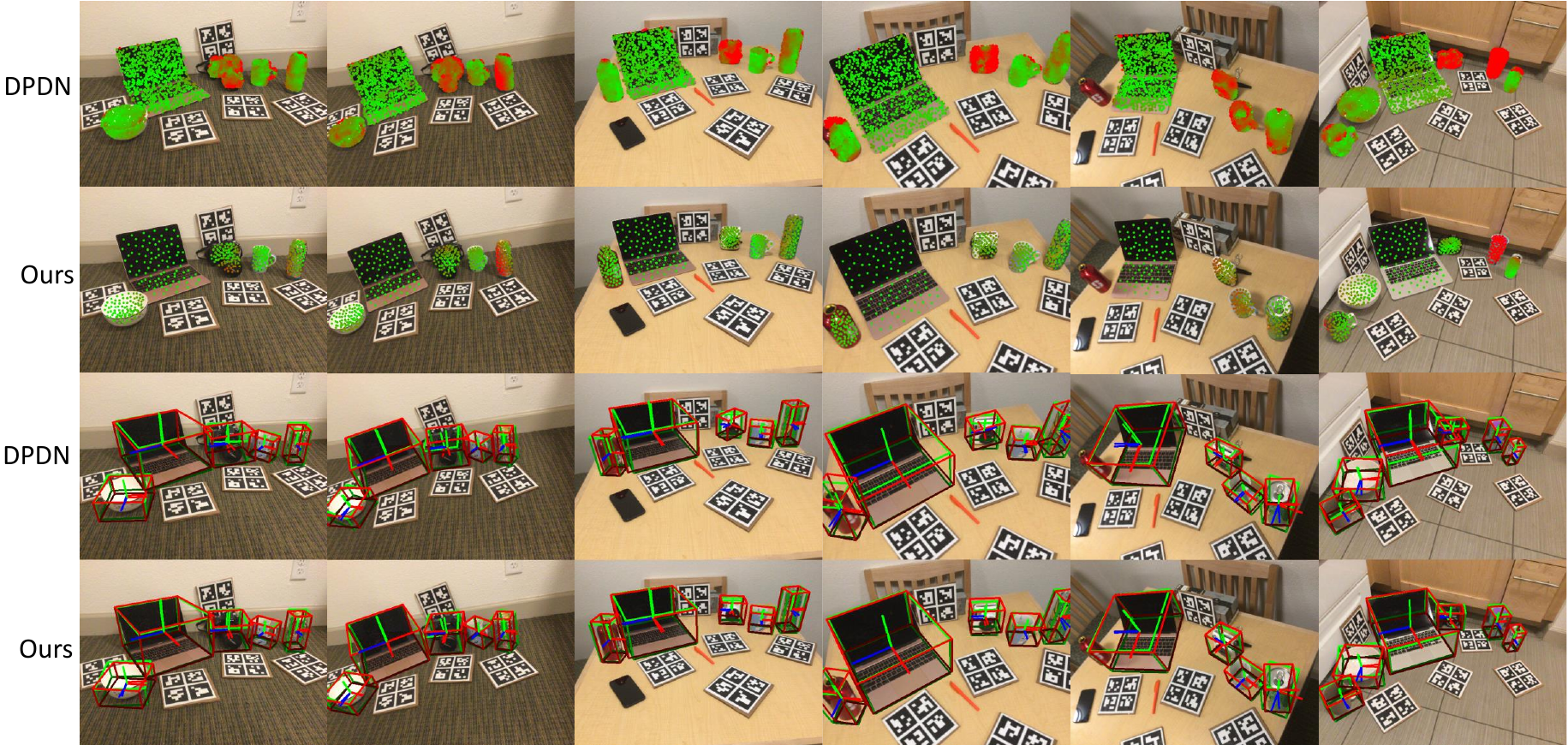}
    \vspace{-0.075in}
    \caption{\textbf{Qualitative comparisons between our method and DPDN \cite{lin2022dpdn} on REAL275 dataset.}
    We visualize the correspondence error maps and pose estimation results of our AG-Pose and DPDN. 
    Red/green indicates large/small errors and predicted/gt bounding boxes. 
    }
    \vspace{-0.35in}
    \label{fig:qualitative_visualizatin}
    \end{center}
\end{figure}

\subsection{Visualization}\label{sec:Visualization}
\textbf{Qualitative Results.}
The qualitative results of DPDN \cite{lin2022dpdn} and 
proposed AG-Pose are shown in Figure \ref{fig:qualitative_visualizatin}. 
Specifically, we visualize the NOCS prediction errors and the final pose predictions
for both methods, in which green/red indicate small/large errors and gt/predicted results.   
The visualization results show that dense correspondence-based methods such as DPDN 
generate larger number of incorrect correspondences on novel instances 
with significant shape variations (e.g., cameras) 
as well as on outlier points (e.g., edge of objects), 
which severely degrade the performance of pose estimation. 
In contrast, our AG-Pose can perform instance-adaptive keypoint detection and
extract geometric-aware features to establish accurate keypoint-level 
correspondences, leading to better pose estimation performance.

\section{Conclusion}
In summary, we present a novel Instance-\textbf{A}daptive and 
\textbf{G}eometric-Aware Keypoint Learning method for category-level 6D object pose estimation (AG-Pose). 
Specifically, we propose a Instance-Adaptive Keypoint Detection module represent the geometric information of different instances through a set of sparse keypoints.  
Futhermore, we propose a Geometric-Aware Feature Aggregation module to 
effectively incorporate local and global geometric information into keypoints to 
establish robust keypoint-level correspondences.
We conduct comprehensive experiments and the experimental results verify the effectiveness of our method.

\section{Acknowledgements}
This work was partially supported by the National Nature Science Foundation of China (Grant 62306294), Dreams Foundation of Jianghuai Advance Technology Center (NO.Q/JH-063-2023-T02A/0) and Anhui Provincial Natural Science Foundation (Grant 2308085QF222).
\newpage
{
    \small
    \bibliographystyle{ieeenat_fullname}
    \bibliography{main}

\begin{thebibliography}{45}
\providecommand{\natexlab}[1]{#1}
\providecommand{\url}[1]{\texttt{#1}}
\expandafter\ifx\csname urlstyle\endcsname\relax
  \providecommand{\doi}[1]{doi: #1}\else
  \providecommand{\doi}{doi: \begingroup \urlstyle{rm}\Url}\fi

\bibitem[Azuma(1997)]{ar3}
Ronald~T Azuma.
\newblock A survey of augmented reality.
\newblock \emph{Presence: teleoperators \& virtual environments}, 6\penalty0 (4):\penalty0 355--385, 1997.

\bibitem[Chen and Dou(2021)]{chen2021sgpa}
Kai Chen and Qi Dou.
\newblock Sgpa: Structure-guided prior adaptation for category-level 6d object pose estimation.
\newblock In \emph{Proceedings of the IEEE/CVF International Conference on Computer Vision}, pages 2773--2782, 2021.

\bibitem[Chen et~al.(2017)Chen, Ma, Wan, Li, and Xia]{ad1}
Xiaozhi Chen, Huimin Ma, Ji Wan, Bo Li, and Tian Xia.
\newblock Multi-view 3d object detection network for autonomous driving.
\newblock In \emph{Proceedings of the IEEE conference on Computer Vision and Pattern Recognition}, pages 1907--1915, 2017.

\bibitem[Di et~al.(2021)Di, Manhardt, Wang, Ji, Navab, and Tombari]{Di2021sopose}
Yan Di, Fabian Manhardt, Gu Wang, Xiangyang Ji, Nassir Navab, and Federico Tombari.
\newblock So-pose: Exploiting self-occlusion for direct 6d pose estimation.
\newblock In \emph{Proceedings of the IEEE/CVF International Conference on Computer Vision (ICCV)}, pages 12396--12405, 2021.

\bibitem[Di et~al.(2022)Di, Zhang, Lou, Manhardt, Ji, Navab, and Tombari]{di2022gpv}
Yan Di, Ruida Zhang, Zhiqiang Lou, Fabian Manhardt, Xiangyang Ji, Nassir Navab, and Federico Tombari.
\newblock Gpv-pose: Category-level object pose estimation via geometry-guided point-wise voting.
\newblock In \emph{Proceedings of the IEEE/CVF Conference on Computer Vision and Pattern Recognition}, pages 6781--6791, 2022.

\bibitem[Duffhauss et~al.(2022)Duffhauss, Demmler, and Neumann]{mv6d}
Fabian Duffhauss, Tobias Demmler, and Gerhard Neumann.
\newblock Mv6d: Multi-view 6d pose estimation on rgb-d frames using a deep point-wise voting network.
\newblock In \emph{2022 IEEE/RSJ International Conference on Intelligent Robots and Systems (IROS)}, pages 3568--3575, 2022.

\bibitem[Geiger et~al.(2012)Geiger, Lenz, and Urtasun]{ad3}
Andreas Geiger, Philip Lenz, and Raquel Urtasun.
\newblock Are we ready for autonomous driving? the kitti vision benchmark suite.
\newblock In \emph{2012 IEEE conference on computer vision and pattern recognition}, pages 3354--3361. IEEE, 2012.

\bibitem[He et~al.(2016)He, Zhang, Ren, and Sun]{he2016resnet}
Kaiming He, Xiangyu Zhang, Shaoqing Ren, and Jian Sun.
\newblock Deep residual learning for image recognition.
\newblock In \emph{Proceedings of the IEEE conference on computer vision and pattern recognition}, pages 770--778, 2016.

\bibitem[He et~al.(2017)He, Gkioxari, Doll{\'a}r, and Girshick]{he2017maskrcnn}
Kaiming He, Georgia Gkioxari, Piotr Doll{\'a}r, and Ross Girshick.
\newblock Mask r-cnn.
\newblock In \emph{Proceedings of the IEEE international conference on computer vision}, pages 2961--2969, 2017.

\bibitem[He et~al.(2020)He, Sun, Huang, Liu, Fan, and Sun]{He2020pvn3d}
Yisheng He, Wei Sun, Haibin Huang, Jianran Liu, Haoqiang Fan, and Jian Sun.
\newblock Pvn3d: A deep point-wise 3d keypoints voting network for 6dof pose estimation.
\newblock In \emph{IEEE/CVF Conference on Computer Vision and Pattern Recognition (CVPR)}, 2020.

\bibitem[He et~al.(2021)He, Huang, Fan, Chen, and Sun]{He2021fffb6d}
Yisheng He, Haibin Huang, Haoqiang Fan, Qifeng Chen, and Jian Sun.
\newblock Ffb6d: A full flow bidirectional fusion network for 6d pose estimation.
\newblock In \emph{IEEE/CVF Conference on Computer Vision and Pattern Recognition (CVPR)}, 2021.

\bibitem[Kingma and Ba(2014)]{kingma2014adam}
Diederik~P Kingma and Jimmy Ba.
\newblock Adam: A method for stochastic optimization.
\newblock \emph{arXiv preprint arXiv:1412.6980}, 2014.

\bibitem[Lee et~al.(2022)Lee, Lee, Shin, Choe, Shin, Kweon, and Yoon]{udacope}
Taeyeop Lee, Byeong-Uk Lee, Inkyu Shin, Jaesung Choe, Ukcheol Shin, In~So Kweon, and Kuk-Jin Yoon.
\newblock Uda-cope: unsupervised domain adaptation for category-level object pose estimation.
\newblock In \emph{Proceedings of the IEEE/CVF Conference on Computer Vision and Pattern Recognition}, pages 14891--14900, 2022.

\bibitem[Lee et~al.(2023)Lee, Tremblay, Blukis, Wen, Lee, Shin, Birchfield, Kweon, and Yoon]{ttacope}
Taeyeop Lee, Jonathan Tremblay, Valts Blukis, Bowen Wen, Byeong-Uk Lee, Inkyu Shin, Stan Birchfield, In~So Kweon, and Kuk-Jin Yoon.
\newblock Tta-cope: Test-time adaptation for category-level object pose estimation.
\newblock In \emph{Proceedings of the IEEE/CVF Conference on Computer Vision and Pattern Recognition}, pages 21285--21295, 2023.

\bibitem[Lin et~al.(2022{\natexlab{a}})Lin, Liu, Cheang, Fu, Guo, and Xue]{lin2022sar}
Haitao Lin, Zichang Liu, Chilam Cheang, Yanwei Fu, Guodong Guo, and Xiangyang Xue.
\newblock Sar-net: Shape alignment and recovery network for category-level 6d object pose and size estimation.
\newblock In \emph{Proceedings of the IEEE/CVF Conference on Computer Vision and Pattern Recognition}, pages 6707--6717, 2022{\natexlab{a}}.

\bibitem[Lin et~al.(2021)Lin, Wei, Li, Xu, Jia, and Li]{dualposenet}
Jiehong Lin, Zewei Wei, Zhihao Li, Songcen Xu, Kui Jia, and Yuanqing Li.
\newblock Dualposenet: Category-level 6d object pose and size estimation using dual pose network with refined learning of pose consistency.
\newblock In \emph{Proceedings of the IEEE/CVF International Conference on Computer Vision (ICCV)}, pages 3560--3569, 2021.

\bibitem[Lin et~al.(2022{\natexlab{b}})Lin, Wei, Ding, and Jia]{lin2022dpdn}
Jiehong Lin, Zewei Wei, Changxing Ding, and Kui Jia.
\newblock Category-level 6d object pose and size estimation using self-supervised deep prior deformation networks.
\newblock In \emph{European Conference on Computer Vision}, pages 19--34. Springer, 2022{\natexlab{b}}.

\bibitem[Liu et~al.(2023)Liu, Chen, Ye, and Qi]{istnet}
Jianhui Liu, Yukang Chen, Xiaoqing Ye, and Xiaojuan Qi.
\newblock Prior-free category-level pose estimation with implicit space transformation.
\newblock \emph{arXiv preprint arXiv:2303.13479}, 2023.

\bibitem[Liu et~al.(2022)Liu, Zhang, Zhang, Fu, Tang, Liang, Tang, Cheng, Zhang, Wang, and Ji]{liu2022gdrnpp_bop}
Xingyu Liu, Ruida Zhang, Chenyangguang Zhang, Bowen Fu, Jiwen Tang, Xiquan Liang, Jingyi Tang, Xiaotian Cheng, Yukang Zhang, Gu Wang, and Xiangyang Ji.
\newblock Gdrnpp.
\newblock \url{https://github.com/shanice-l/gdrnpp_bop2022}, 2022.

\bibitem[Manhardt et~al.(2019)Manhardt, Kehl, and Gaidon]{ad2}
Fabian Manhardt, Wadim Kehl, and Adrien Gaidon.
\newblock Roi-10d: Monocular lifting of 2d detection to 6d pose and metric shape.
\newblock In \emph{Proceedings of the IEEE/CVF Conference on Computer Vision and Pattern Recognition}, pages 2069--2078, 2019.

\bibitem[Marchand et~al.(2015)Marchand, Uchiyama, and Spindler]{ar1}
Eric Marchand, Hideaki Uchiyama, and Fabien Spindler.
\newblock Pose estimation for augmented reality: A hands-on survey.
\newblock \emph{IEEE Transactions on Visualization and Computer Graphics (TVCG)}, 22\penalty0 (12):\penalty0 2633--2651, 2015.

\bibitem[Mo et~al.(2022)Mo, Gan, Yokoya, and Chen]{mo2022es6d}
Ningkai Mo, Wanshui Gan, Naoto Yokoya, and Shifeng Chen.
\newblock Es6d: A computation efficient and symmetry-aware 6d pose regression framework.
\newblock In \emph{Proceedings of the IEEE/CVF Conference on Computer Vision and Pattern Recognition}, pages 6718--6727, 2022.

\bibitem[Mousavian et~al.(2019)Mousavian, Eppner, and Fox]{robotmani2}
Arsalan Mousavian, Clemens Eppner, and Dieter Fox.
\newblock 6-dof graspnet: Variational grasp generation for object manipulation.
\newblock In \emph{Proceedings of the IEEE/CVF International Conference on Computer Vision}, pages 2901--2910, 2019.

\bibitem[Park et~al.(2019)Park, Patten, and Vincze]{Park2019pix2pose}
Kiru Park, Timothy Patten, and Markus Vincze.
\newblock Pix2pose: Pix2pose: Pixel-wise coordinate regression of objects for 6d pose estimation.
\newblock In \emph{The IEEE International Conference on Computer Vision (ICCV)}, 2019.

\bibitem[Peng et~al.(2019)Peng, Liu, Huang, Zhou, and Bao]{peng2019pvnet}
Sida Peng, Yuan Liu, Qixing Huang, Xiaowei Zhou, and Hujun Bao.
\newblock Pvnet: Pixel-wise voting network for 6dof pose estimation.
\newblock In \emph{IEEE/CVF Conference on Computer Vision and Pattern Recognition (CVPR)}, 2019.

\bibitem[Qi et~al.(2017)Qi, Yi, Su, and Guibas]{qi2017pointnet++}
Charles~Ruizhongtai Qi, Li Yi, Hao Su, and Leonidas~J Guibas.
\newblock Pointnet++: Deep hierarchical feature learning on point sets in a metric space.
\newblock In \emph{Advances in Neural Information Processing Systems (NeurIPS)}, pages 5099--5108, 2017.

\bibitem[Smith(2017)]{smith2017cyclical}
Leslie~N Smith.
\newblock Cyclical learning rates for training neural networks.
\newblock In \emph{2017 IEEE winter conference on applications of computer vision (WACV)}, pages 464--472. IEEE, 2017.

\bibitem[Tian et~al.(2020)Tian, Ang~Jr, and Lee]{Tian2020spd}
Meng Tian, Marcelo~H Ang~Jr, and Gim~Hee Lee.
\newblock Shape prior deformation for categorical 6d object pose and size estimation.
\newblock In \emph{Proceedings of the European Conference on Computer Vision (ECCV)}, 2020.

\bibitem[Umeyama(1991)]{umeyama1991least}
Shinji Umeyama.
\newblock Least-squares estimation of transformation parameters between two point patterns.
\newblock \emph{IEEE Transactions on Pattern Analysis \& Machine Intelligence}, 13\penalty0 (04):\penalty0 376--380, 1991.

\bibitem[Vaswani et~al.(2017)Vaswani, Shazeer, Parmar, Uszkoreit, Jones, Gomez, Kaiser, and Polosukhin]{vaswani2017attention}
Ashish Vaswani, Noam Shazeer, Niki Parmar, Jakob Uszkoreit, Llion Jones, Aidan~N Gomez, {\L}ukasz Kaiser, and Illia Polosukhin.
\newblock Attention is all you need.
\newblock \emph{Advances in neural information processing systems}, 30, 2017.

\bibitem[Wang et~al.(2019{\natexlab{a}})Wang, Xu, Zhu, Mart{\'\i}n-Mart{\'\i}n, Lu, Fei-Fei, and Savarese]{wang2019densefusion}
Chen Wang, Danfei Xu, Yuke Zhu, Roberto Mart{\'\i}n-Mart{\'\i}n, Cewu Lu, Li Fei-Fei, and Silvio Savarese.
\newblock Densefusion: 6d object pose estimation by iterative dense fusion.
\newblock In \emph{Proceedings of the IEEE/CVF conference on computer vision and pattern recognition}, pages 3343--3352, 2019{\natexlab{a}}.

\bibitem[Wang et~al.(2021{\natexlab{a}})Wang, Manhardt, Tombari, and Ji]{Wang2021GDRN}
Gu Wang, Fabian Manhardt, Federico Tombari, and Xiangyang Ji.
\newblock {GDR-Net}: Geometry-guided direct regression network for monocular 6d object pose estimation.
\newblock In \emph{IEEE/CVF Conference on Computer Vision and Pattern Recognition (CVPR)}, pages 16611--16621, 2021{\natexlab{a}}.

\bibitem[Wang et~al.(2019{\natexlab{b}})Wang, Sridhar, Huang, Valentin, Song, and Guibas]{Wang2019nocs}
He Wang, Srinath Sridhar, Jingwei Huang, Julien Valentin, Shuran Song, and Leonidas~J. Guibas.
\newblock Normalized object coordinate space for category-level 6d object pose and size estimation.
\newblock In \emph{The IEEE Conference on Computer Vision and Pattern Recognition (CVPR)}, 2019{\natexlab{b}}.

\bibitem[Wang et~al.(2021{\natexlab{b}})Wang, Chen, and Dou]{wang2021care}
Jiaze Wang, Kai Chen, and Qi Dou.
\newblock Category-level 6d object pose estimation via cascaded relation and recurrent reconstruction networks.
\newblock In \emph{2021 IEEE/RSJ International Conference on Intelligent Robots and Systems (IROS)}, pages 4807--4814. IEEE, 2021{\natexlab{b}}.

\bibitem[Wang et~al.(2023)Wang, Wang, Li, Yang, Wan, and Liu]{query6dof}
Ruiqi Wang, Xinggang Wang, Te Li, Rong Yang, Minhong Wan, and Wenyu Liu.
\newblock Query6dof: Learning sparse queries as implicit shape prior for category-level 6dof pose estimation.
\newblock In \emph{Proceedings of the IEEE/CVF International Conference on Computer Vision}, pages 14055--14064, 2023.

\bibitem[Wen et~al.(2022)Wen, Lian, Bekris, and Schaal]{robotmani3}
Bowen Wen, Wenzhao Lian, Kostas Bekris, and Stefan Schaal.
\newblock You only demonstrate once: Category-level manipulation from single visual demonstration.
\newblock \emph{Robotics: Science and Systems (RSS)}, 2022.

\bibitem[Wu et~al.(2020)Wu, Chen, Cao, Zhang, Tai, Sun, and Jia]{robotmani5}
Chaozheng Wu, Jian Chen, Qiaoyu Cao, Jianchi Zhang, Yunxin Tai, Lin Sun, and Kui Jia.
\newblock Grasp proposal networks: An end-to-end solution for visual learning of robotic grasps.
\newblock \emph{Advances in Neural Information Processing Systems}, 33:\penalty0 13174--13184, 2020.

\bibitem[Wu et~al.(2022{\natexlab{a}})Wu, Javaheri, Zand, and Greenspan]{wu2022keypointrefine}
Yangzheng Wu, Alireza Javaheri, Mohsen Zand, and Michael Greenspan.
\newblock Keypoint cascade voting for point cloud based 6dof pose estimation.
\newblock In \emph{2022 International Conference on 3D Vision (3DV)}. IEEE, 2022{\natexlab{a}}.

\bibitem[Wu et~al.(2022{\natexlab{b}})Wu, Zand, Etemad, and Greenspan]{wu2022vote}
Yangzheng Wu, Mohsen Zand, Ali Etemad, and Michael Greenspan.
\newblock Vote from the center: 6 dof pose estimation in rgb-d images by radial keypoint voting.
\newblock In \emph{European Conference on Computer Vision (ECCV)}. Springer, 2022{\natexlab{b}}.

\bibitem[Xiang et~al.(2017)Xiang, Schmidt, Narayanan, and Fox]{xiang2018posecnn}
Yu Xiang, Tanner Schmidt, Venkatraman Narayanan, and Dieter Fox.
\newblock Posecnn: A convolutional neural network for 6d object pose estimation in cluttered scenes.
\newblock \emph{arXiv preprint arXiv:1711.00199}, 2017.

\bibitem[Xu et~al.(2022)Xu, Lin, Zhang, Wang, and Li]{xu2022rnnpose}
Yan Xu, Kwan-Yee Lin, Guofeng Zhang, Xiaogang Wang, and Hongsheng Li.
\newblock Rnnpose: Recurrent 6-dof object pose refinement with robust correspondence field estimation and pose optimization.
\newblock In \emph{Proceedings of the IEEE/CVF Conference on Computer Vision and Pattern Recognition}, 2022.

\bibitem[Zhang et~al.(2022)Zhang, Di, Lou, Manhardt, Tombari, and Ji]{zhang2022rbp}
Ruida Zhang, Yan Di, Zhiqiang Lou, Fabian Manhardt, Federico Tombari, and Xiangyang Ji.
\newblock Rbp-pose: Residual bounding box projection for category-level pose estimation.
\newblock In \emph{European Conference on Computer Vision}, pages 655--672. Springer, 2022.

\bibitem[Zhao et~al.(2017)Zhao, Shi, Qi, Wang, and Jia]{zhao2017psp}
Hengshuang Zhao, Jianping Shi, Xiaojuan Qi, Xiaogang Wang, and Jiaya Jia.
\newblock Pyramid scene parsing network.
\newblock In \emph{Proceedings of the IEEE conference on computer vision and pattern recognition}, pages 2881--2890, 2017.

\bibitem[Zheng et~al.(2023)Zheng, Wang, Sun, Dasgupta, Chen, Leonardis, Zhang, and Chang]{Zheng2023hspose}
Linfang Zheng, Chen Wang, Yinghan Sun, Esha Dasgupta, Hua Chen, Ale\v{s} Leonardis, Wei Zhang, and Hyung~Jin Chang.
\newblock Hs-pose: Hybrid scope feature extraction for category-level object pose estimation.
\newblock In \emph{Proceedings of the IEEE/CVF Conference on Computer Vision and Pattern Recognition (CVPR)}, pages 17163--17173, 2023.

\bibitem[Zhou et~al.(2019)Zhou, Barnes, Lu, Yang, and Li]{zhou2019continuity6d}
Yi Zhou, Connelly Barnes, Jingwan Lu, Jimei Yang, and Hao Li.
\newblock On the continuity of rotation representations in neural networks.
\newblock In \emph{Proceedings of the IEEE/CVF Conference on Computer Vision and Pattern Recognition}, pages 5745--5753, 2019.

\end{thebibliography}
}


\end{document}